%% file: main.tex
\documentclass[runningheads]{llncs}

\usepackage{eccvabbrv}
\usepackage{microtype}
\usepackage{graphicx}
\usepackage{subcaption}
\usepackage{booktabs}

\usepackage[accsupp]{axessibility}

\usepackage{hyperref}

\usepackage{amsmath}
\usepackage{amssymb}
\usepackage{mathtools}
\usepackage{makecell}
\usepackage{multirow}
\usepackage{bm}

\usepackage{algorithm}
\usepackage{algorithmic}

\newcommand{\systemname}{DeCoR}
\newcommand{\papertitle}{\systemname: Design and Control Co-Optimization for Urban Streets Using Reinforcement Learning}

\title{\papertitle}
\titlerunning{\systemname: Design and Control Co-Optimization}

\author{Bibek Poudel\inst{1} \and Lei Zhu\inst{2} \and Kevin Heaslip\inst{1} \and Sai Swaminathan\inst{1} \and Weizi Li\inst{4}}
\authorrunning{B. Poudel et al.}
\institute{University of Tennessee, Knoxville, TN, USA \and University of North Carolina at Charlotte, Charlotte, NC, USA \and University of California, Riverside, CA, USA}

\begin{document}

\maketitle

\input{sections/abstract}

\input{sections/intro}

\input{sections/related}

\input{sections/method}

\input{sections/experiments}

\input{sections/conclusion}
\input{sections/acknowledgements}

\bibliographystyle{splncs04}
\bibliography{main}

\clearpage
\setcounter{page}{1}
\appendix
\input{sections/supplementary}


\end{document}

%% file: sections/abstract.tex
\begin{abstract}

Modern vision systems can detect, track, and forecast urban actors at scale, yet translating perception outputs to urban design remains limited. We introduce \systemname{}, a two-stage reinforcement learning framework that leverages flow observations to co-optimize crosswalk layout and network-level signal control. The design stage encodes the pedestrian network as a graph and learns a generative policy that parameterizes a Gaussian mixture model over crosswalk location and width, from which new crosswalks are sampled. For each layout, a shared control policy learns adaptive signal timings to minimize joint pedestrian and vehicle delay. On a 750\,m real-world urban corridor with demand sensed from video and Wi-Fi logs, \systemname{} learns a layout that reduces pedestrian arrival time to their nearest crosswalk by 23\% while using fewer crosswalks than existing configurations. On the control side, \systemname{} reduces pedestrian and vehicle wait time by 79\% and 65\%, respectively, relative to fixed-time signalization. Further, the control policy generalizes to demands outside of training and is robust to layout changes without retraining. Our code and data are publicly available in \url{https://github.com/poudel-bibek/DeCoR}.

\end{abstract}

%% file: sections/intro.tex
\section{Introduction}
\label{sec:intro}

Modern computer vision has made remarkable progress in perceiving urban scenes, from identifying and tracking pedestrians~\cite{kong2024wts}, to predicting their trajectories~\cite{mohamed2020social}, generating city layouts~\cite{xie2024citydreamer,he2024coho}, and building simulation platforms for autonomous fleet~\cite{tan2025scenediffuser++,suo2021trafficsim,guo2024lasil}. Despite these advances, the step from \emph{perception to design} remains largely unaddressed: we can increasingly measure how pedestrians and vehicles move through cities, but we rarely use those measurements to design the infrastructure that shapes their behavior. This gap is critical because pedestrian fatalities continue to rise~\cite{marshall2024killed}. In the United States alone, \(7{,}508\) pedestrians were killed in \(2022\), the highest toll in \(40\) years, with \(84\%\) of the fatalities on urban roads mostly at mid-block locations without any crossing infrastructure~\cite{GHSA2023,NHTSA2023,NSC2023}.



Established guidance on pedestrian safety emphasizes that infrastructure design, including crosswalk placement, strongly shapes safety outcomes~\cite{chandler2013signalized,FHWA2021}. Yet current practice relies heavily on heuristics such as crosswalk spacing thresholds and minimum distances from intersections,
many of which predate modern pedestrian and vehicle sensing capabilities~\cite{traffic2023manual,ITEcrosswalkpolicy2022,blackburn2018guide,marshall2024killed}. Further, these rules often leave practitioners to guesswork when balancing competing stakeholder priorities, as pedestrians want safe and convenient crossings, vehicles expect minimal stops and delays, and planners aim to reduce costs and maintain efficiency.


On urban streets, which often carry high pedestrian volume, balancing these priorities hinges on two coupled decisions: \emph{where crosswalks are placed} and \emph{how they are regulated}. Placement shapes pedestrian route choice and vehicle-pedestrian conflict points, while control determines the timing of how those conflicts are resolved. However, optimizing one without the other can be counterproductive. Adding crosswalks may shorten walking detours, but if they are not aligned with pedestrian desire lines they unnecessarily increase vehicle stops and reduce capacity. Conversely, a crosswalk well-aligned with desire lines but paired with poor signal timing can cause excessive delays. 
In turn, even well-timed signals on a poorly designed layout struggle to compensate for missing crossings where pedestrians need them. 
This coupling motivates a perception-driven co-optimization approach that jointly addresses design and control~\cite{cong2015co,jha2006intelligent}.


\begin{figure*}[t!]
    \centering
    \includegraphics[width=1.0\linewidth]{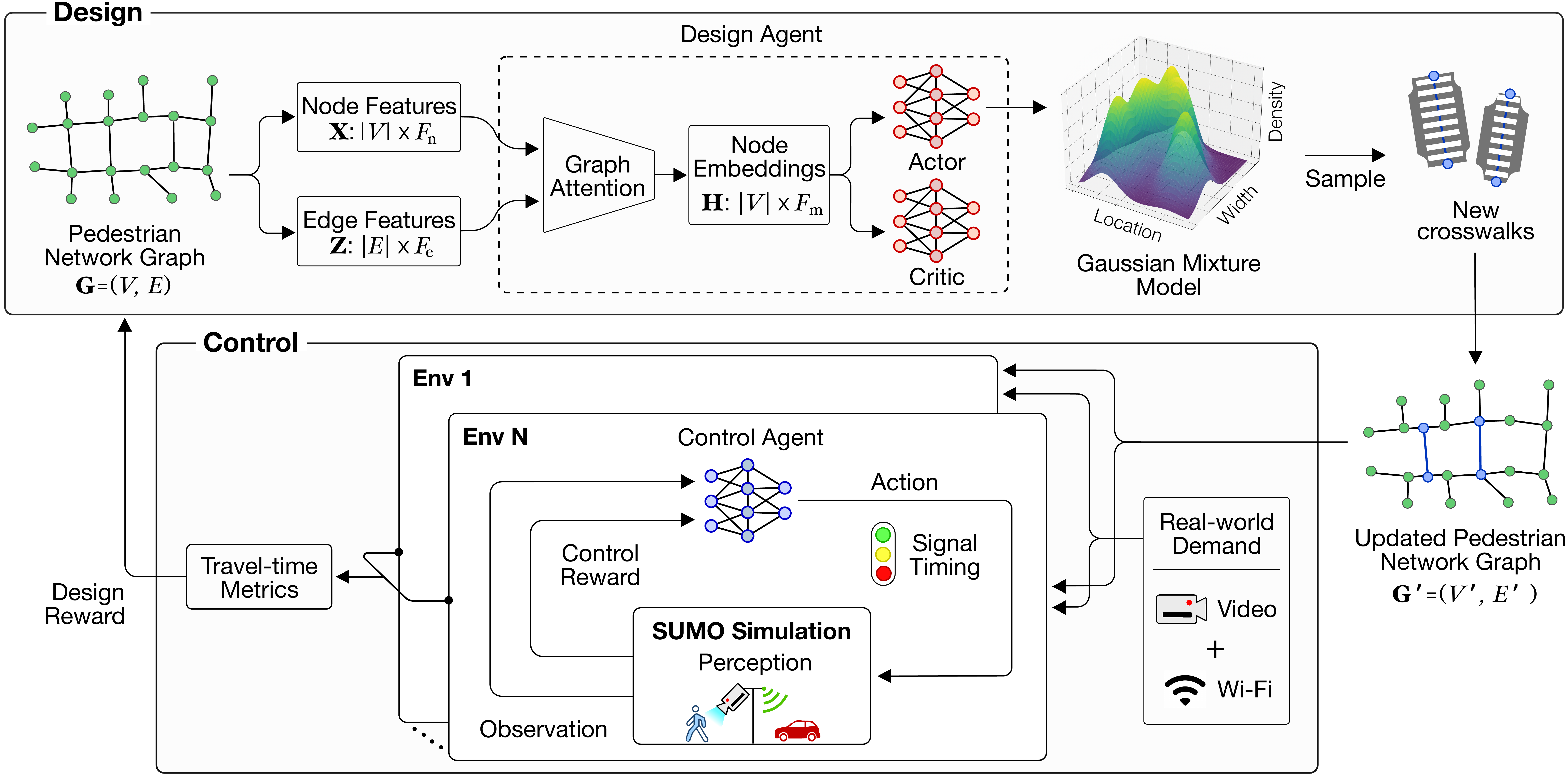}
        \caption{The two-stage co-optimization loop in \systemname{}. \textbf{Design:} A graph-attention encoder takes the pedestrian-network graph \(G\) as input and parameterizes a GMM over crosswalk location and width. New crosswalks (blue) are sampled from the GMM each episode and integrated into \(G\), producing the updated layout \(G'\). \textbf{Control:} \(G'\) is evaluated across \(N\) parallel closed-loop simulations, each with randomly scaled real-world demand. At each step, observations for the control agent are obtained from perceived pedestrian and vehicle states, from which a single shared control policy determines signal timings. Upon episode end, travel time metrics are averaged across the \(N\) environments to compute the control and design rewards.}
    \vspace{-12pt}
    \label{fig:system}
\end{figure*}


We present \systemname{}, a two-stage reinforcement learning (RL) framework that co-optimizes mid-block crosswalk layout and network-level signal control, shown in Fig.~\ref{fig:system}. \systemname{} addresses perception to design by evaluating each proposed layout through closed-loop simulation, as the quality of a placement depends on how pedestrians and vehicles interact with it. Perception drives this evaluation at two timescales. Offline, real-world pedestrian and vehicle demand sensed from video and Wi-Fi logs grounds the simulation in observed behavior. Online, a control policy uses per-step observations of pedestrian and vehicle states to adapt signal timings at every step. At each episode end, the resulting pedestrian and vehicle travel time metrics provide learning signals for two complementary policies. The design policy encodes the pedestrian network as a graph and learns a generative model over crosswalk location and width, from which new crossings are sampled, while the control policy optimizes signal timings for each sampled layout. We evaluate \systemname{} on a 750\,m corridor within a university campus in the U.S. with \(2{,}223\) pedestrians and \(202\) vehicles per hour. The policies from \systemname{} substantially reduce travel times for both pedestrians and vehicles compared to the real-world layout and fixed-time signal plans, demonstrating that co-optimization can outperform conventional placement heuristics and traditional control. Our contributions are as follows.  
\begin{enumerate}
    \item We introduce \systemname{}, a perception-driven framework that jointly optimizes crosswalk layout and traffic signal control through a two-stage RL formulation, with a generative design agent and a network-level adaptive controller.
    \item The design agent discovers crosswalk configurations that reduce pedestrian arrival times by up to 23\% compared to the real-world layout, despite using fewer crosswalks. Given this layout, the control agent reduces pedestrian and vehicle waiting time by 79\% and 65\%, respectively, relative to fixed-time signals, closely matching the performance of unsignalized crossings.
    \item Both agents generalize to demand patterns and scales outside the training range. Further, the control agent trained under co-optimization remains robust when an additional crosswalk is introduced to the optimized layout, achieving up to 97\% lower wait times than a sequentially trained agent.
\end{enumerate}
\vspace{-2pt}
Our demo videos, code, and data are available in the supplementary material.

%% file: sections/related.tex
\section{Related Work}
\label{sec:related}

Recent progress in computer vision has substantially improved our ability to perceive pedestrian and vehicle behavior in urban scenes. Pedestrian-centric datasets and scene understanding benchmarks enable fine-grained spatiotemporal reasoning from traffic videos~\cite{kong2024wts}, and graph-based models now capture social interactions in trajectory forecasting~\cite{mohamed2020social,yu2020spatio} as well as heterogeneous vehicle--pedestrian dynamics~\cite{mo2022multi,diehl2019graph}. In parallel, learned multi-agent simulators and generative world models provide increasingly realistic evaluation of multi-agent behavior under a fixed infrastructure~\cite{suo2021trafficsim,guo2024lasil,tan2025scenediffuser++}, with recent work addressing safety-critical settings~\cite{chang2024safe}, closed-loop traffic generation~\cite{lin2025causal,zhang2025closed}, and city-scale layout synthesis for visual realism~\cite{xie2024citydreamer,he2024coho}. Additionally, Wi-Fi sensing complements these efforts with real-world demand estimation, adding further realism~\cite{zhang2023crowdtelescope,yuan2025hierarchical}. Nevertheless, these advances remain confined to sensing, prediction, and simulation of behavior under a given infrastructure; the inverse step from perception to design remains comparatively underexplored.

Reinforcement learning (RL) has begun to close this gap in adjacent spatial design problems, including land-use planning~\cite{zheng2023spatial}, road layout for informal settlements~\cite{Zheng2023SlumRoad}, and adaptive road configurations~\cite{ye2022adaptive}, though without jointly optimizing infrastructure with its downstream control. Separately, traffic signal control, though extensively studied with RL through multi-agent formulations~\cite{wiering2000multi,ElTantawy2013,chen2020toward,Xu2021HiLight} and graph- and attention-based architectures~\cite{Nishi2018GCNsignal,wei2019colight,oroojlooy2020attendlight,wu2023transformerlight}, takes the layout as given and remains predominantly vehicle-centric; only a few formulations incorporate pedestrian delay~\cite{Yazdani2023IVPL,poudel2025control}. Recent co-optimization efforts have jointly tuned vehicular flow directions~\cite{zhao2023deep} and freeway network topology~\cite{cong2015co} with signal control, but target vehicle-only objectives. Across these three lines of work, methods either optimize layout without accounting for operational control, or co-optimize design and control for vehicular objectives alone, and most are not grounded in real-world network or demand. Our work addresses these limitations by jointly optimizing pedestrian-focused crosswalk design and network-level signal control, with both network and demand derived from the real world.

%% file: sections/method.tex
\section{Methodology}
\label{sec:method}

As illustrated in Fig.~\ref{fig:system}, \systemname{} operates in two stages---an outer design stage that proposes a crosswalk layout each round and an inner control stage that learns network-level signal timings conditioned on each proposed layout. The underlying network is represented as a directed graph $G = (V, E)$, where nodes $V$ correspond to intersections, crosswalk endpoints, and path junctions, and edges $E$ correspond to walkable segments connecting them. Nodes carry coordinate features $x_v \in \mathbb{R}^{F_n}$ and edges carry length and width features $z_e \in \mathbb{R}^{F_e}$. The graph contains two types of controlled locations: intersections, whose geometry is fixed, and mid-block crosswalks, signalized pedestrian crossings whose placement and width are proposed by the design policy. The control policy then sets signal timings across both. For brevity, we refer to fixed intersections as ``intersections'' and designed mid-block crossings as ``crosswalks,'' collectively as ``signals.''

We jointly optimize the design policy $\pi_D$ and control policy $\pi_C$, parameterized by $\theta_D$ and $\theta_C$, via reinforcement learning:
\vspace{-6pt}
\begin{align}
\theta_D^\star &= \underset{\theta_D}{\arg\max}
  \Bigl\{
    \mathbb{E}_{\mathcal{C}^{(k)} \sim \pi_D(\cdot|G^{(k)})}
    \bigl[
      R_D(G^{(k+1)})
    \bigr]
  \Bigr\}, \label{eq:design_reward}\\[6pt]
\theta_C^\star &= \underset{\theta_C}{\arg\max}
  \Bigl\{
    \mathbb{E}_{\tau\sim\pi_C(\cdot|G^{(k+1)})}
    \bigl[
      \textstyle\sum_{t=0}^{T-1}\gamma_C^{t}\,r_{C,t}(s_t,a_{C,t};G^{(k+1)})
    \bigr]
  \Bigr\}, \label{eq:control_reward}
\end{align}
where \(G^{(k+1)} = G^{\text{base}} \cup \mathcal{C}^{(k)}\), subject to \(|\mathcal{C}^{(k)}|\le N_{\max}\), \(\forall c\in\mathcal{C}^{(k)},\; \text{location}(c)\in[l_{\min}, l_{\max}]\), and \(\text{width}(c)\in[w_{\min}, w_{\max}]\). Here, $G^{\text{base}}$ is the initial graph without crosswalks. At each round $k$, the design policy proposes up to $N_{\max}$ crosswalks $\mathcal{C}^{(k)}$ that are inserted into $G^{\text{base}}$ to form $G^{(k+1)}$, with each crosswalk's location and width bounded by corridor limits. The design reward $R_D$ balances pedestrian arrival times against the number of crosswalks, capturing the trade-off between convenience and infrastructure cost. Conditioned on $G^{(k+1)}$, the control policy $\pi_C$ runs for $T$ steps in each of $N$ parallel simulations. In each simulation $i$, the real-world demand $\mathcal{D}$ is scaled by a random factor $\alpha_i\sim\mathcal{U}[\alpha_{\min},\alpha_{\max}]$ to obtain $\mathcal{D}_i=\alpha_i\mathcal{D}$, exposing the policy to demand variation. At each step $t$, the policy perceives pedestrian and vehicle state $s_t$, selects actions $a_{C,t}$, and receives instantaneous reward $r_{C,t}$ based on combined pedestrian and vehicle delays, discounted by $\gamma_C$. The complete procedure is given in Algorithm~\ref{alg:design_control}, with $G^{(0)} = G^{\text{base}} \cup \mathcal{C}^{(0)}$ representing the original real-world network layout. Because each round rebuilds the layout from $G^{\text{base}}$, the design stage is a contextual bandit with an immediate reward, while the control stage is a Markov Decision Process (MDP) with sequential state transitions over $T$ steps.
\setlength{\textfloatsep}{8pt plus 2pt minus 2pt}
\begin{algorithm}[t!]
\caption{\systemname{}: Design and Control Co-optimization}
\label{alg:design_control}
\small
\begin{algorithmic}[1]
\STATE \textbf{Input:} graph $G^{(0)}=(V^{(0)},E^{(0)})$, demand $\mathcal{D}$
\STATE \textbf{Initialize:} policies $(\pi_{D}~;~\theta_{D})$, $(\pi_{C}~;~\theta_{C})$, buffers: design $\mathcal{B}_{D}$, control $\mathcal{B}_{C}$
\FOR{co-optimization \textbf{round} $k = 0,1,\dots,K-1$}
    \STATE Extract node features $X^{(k)}\in\mathbb{R}^{|V^{(k)}|\times F_{n}}$ and edge features $Z^{(k)}\in\mathbb{R}^{|E^{(k)}|\times F_{e}}$
    \STATE $H^{(k)} \gets \text{GraphAttention}\bigl(X^{(k)},Z^{(k)};\theta_{D}\bigr)$
    \STATE $a_{D}^{(k)} \sim \pi_{D}\!\bigl(H^{(k)};\theta_{D}\bigr)$
    \STATE Sample crosswalks $\mathcal{C}^{(k)} \sim \text{GMM}\bigl(a_{D}^{(k)}\bigr)$
    \STATE $G^{(k+1)} \gets G^{\text{base}} \cup \mathcal{C}^{(k)}$
    \STATE Launch $N$ parallel envs $\{\text{Env}_{i}\}_{i=1}^{N}$ from $G^{(k+1)}$ with shared $\pi_{C}$
    \FORALL{$i = 1,\dots,N$ \textbf{in parallel}}
        \STATE Sample $\alpha_i \sim \mathcal{U}[\alpha_{\min},\alpha_{\max}]$; initialize $\text{Env}_i$ with $\mathcal{D}_i = \alpha_i\,\mathcal{D}$
        \FOR{$t = 0$ \textbf{to} $T-1$}
            \STATE Observe state $s_{t}^{i}$ from $\text{Env}_{i}$
            \STATE $a_{C,t}^{i} \sim \pi_{C}\!\bigl(s_{t}^{i};\theta_{C}\bigr)$
            \STATE Execute $a_{C,t}^{i}$, obtain $s_{t+1}^{i}$ and $r_{C,t}^{i}$
            \STATE Store $(s_{t}^{i},a_{C,t}^{i},s_{t+1}^{i},r_{C,t}^{i})$ in $\mathcal{B}_{C}$
        \ENDFOR
        \STATE Compute design reward $R_{D,i}^{(k)}$ from $\text{Env}_{i}$
    \ENDFOR
    \STATE $R_{D}^{(k)} \gets \dfrac{1}{N}\sum_{i=1}^{N} R_{D,i}^{(k)}$
    \STATE $\theta_{C} \gets \text{PPOUpdate}\bigl(\pi_{C},\mathcal{B}_{C}\bigr)$
    \STATE Store $\bigl(G^{(k)},\mathcal{C}^{(k)},G^{(k+1)},R_{D}^{(k)}\bigr)$ in $\mathcal{B}_{D}$
    \STATE $\theta_{D} \gets \text{PPOUpdate}\bigl(\pi_{D},\mathcal{B}_{D}\bigr)$
\ENDFOR
\end{algorithmic}
\end{algorithm}



Each round, the policy observes $G^{(k)}$ and proposes crosswalks $\mathcal{C}^{(k)}$ that are inserted into $G^{\text{base}}$; the resulting layout is scored via closed-loop simulation in SUMO~\cite{krajzewicz2002sumo}. Both stages are optimized with Proximal Policy Optimization~\cite{schulman2017proximal}; its stochastic exploration and advantage-based variance reduction~\cite{schulman2015high} suit the continuous design space and stabilize learning across $N$ parallel simulations, each with heterogeneous demand. The design bandit is defined as follows.
\begin{itemize}
    \item \textbf{Context.} The pedestrian network graph $G^{(k)} = G^{\text{base}} \cup \mathcal{C}^{(k-1)}$ at round $k$, represented by node features $X^{(k)}$ encoding normalized coordinates and edge features $Z^{(k)}$ encoding normalized lengths and widths.
    \item \textbf{Action.} The action $a_D^{(k)}$ specifies the means $(\mu_m)$ of an $M=7$ component Gaussian Mixture Model (GMM) over crosswalk location and width, with probability density:
    \begin{equation}
    p(x) = \frac{1}{M}\sum_{m=1}^{M}\mathcal{N}\left(x\,\middle|\,\mu_m,\,\sigma^2 I\right),
    \label{eq:gmm}
    \vspace{-6pt}
    \end{equation}
    where $\sigma^2$ is the variance and $I$ is the identity matrix. The diagonal covariance matrix ($\sigma^2 I$) and mixture weights are kept fixed, so only the means are learned. During training, we sample crosswalk proposals $\mathcal{C}^{(k)}$ stochastically from this GMM to encourage exploration; during evaluation, we deterministically extract the local maxima of $p(x)$ to produce consistent designs. Further details are in the supplementary material~\ref{appen:policy_gmm}.
    
    \item \textbf{Reward.} The design reward $R_D^{(k)}$ captures the trade-off between pedestrian convenience and infrastructure cost, averaged across $N$ parallel simulations with varying demand:
    \begin{equation}
    R_D^{(k)} = \frac{1}{N}\sum_{i=1}^{N}\Bigl(\lambda_{\text{1}} \frac{1}{|\mathcal{P}|}\sum_{p \in \mathcal{P}} t_{\text{arrival},p} + \lambda_{\text{2}} |\mathcal{C}^{(k)}|\Bigr),
    \label{eq:reward_function}
    \vspace{-6pt}
    \end{equation}
    
    where $\mathcal{P}$ denotes the set of simulated pedestrians and $t_{\text{arrival},p}$ is the arrival time of pedestrian $p$. We set $\lambda_{\text{1}}=-1$ to penalize larger arrival times and $\lambda_{\text{2}}=-2$ to encourage sparser, cost-effective designs.
\end{itemize}

\begin{figure*}[t!]
    \centering
    \includegraphics[width=0.90\linewidth]{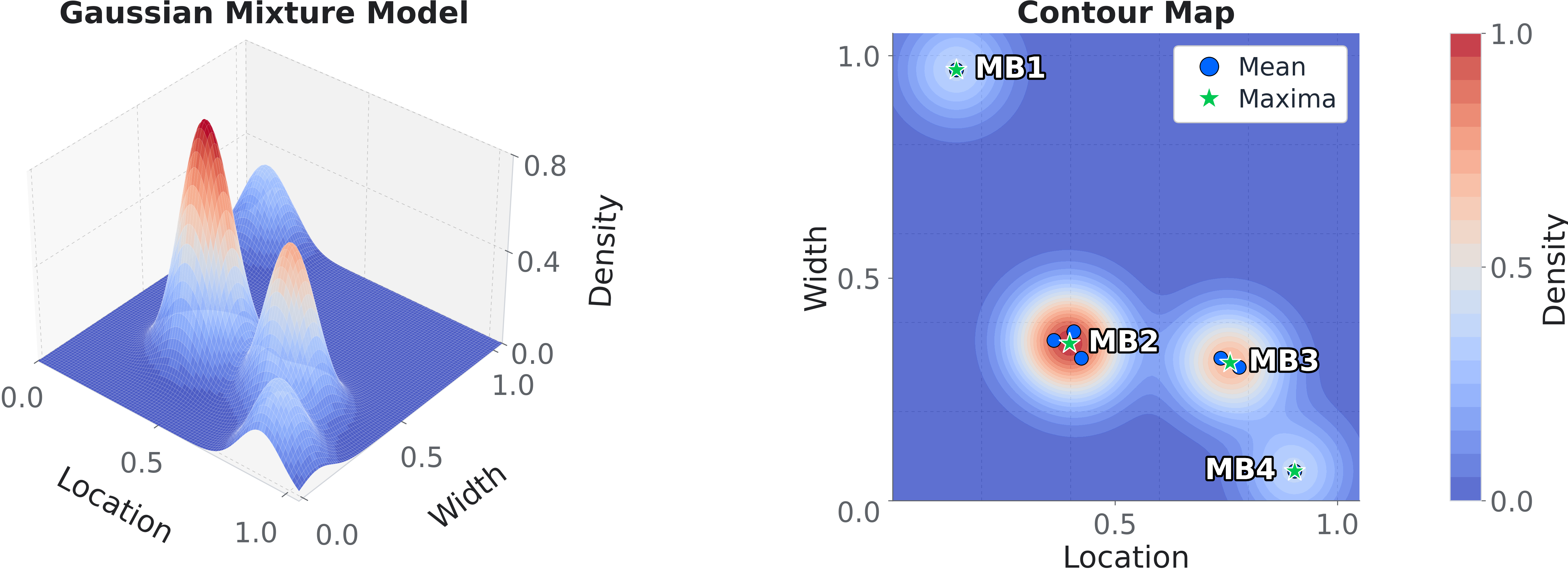}
    \vspace{-2pt}
    \caption{\textbf{LEFT:} Learned Gaussian mixture model (GMM) over normalized crosswalk location and width, with modes corresponding to preferred configurations. \textbf{RIGHT:} Top-down view of the GMM with seven component means and four local maxima of widths \(12\)\,m, \(6\)\,m, \(7\)\,m, and \(2\)\,m, respectively. Although the GMM has seven components, only four mid-block crosswalks (MB\(1\)--\(4\)) are obtained as multiple means collapse to a single maximum: three at MB\(2\) and two at MB\(3\).}

    \label{fig:network_gmm}
    \vspace{6pt}
\end{figure*}

Because the reward penalizes crosswalk count, the GMM can exhibit mode collapse, where several Gaussian components converge to nearly the same location, forming a single dominant peak of the density function and yielding only one crosswalk proposal despite originating from multiple mixture components, as illustrated in Fig.~\ref{fig:network_gmm}. Separately, to rule out physically unrealistic configurations in which two crosswalks lie closer than $1\;\text{m}$, we merge every such pair, replacing them with a single proposal at the mean of their locations and widths. Although we derive the advantage from the reward of the merged proposals, to ensure correct credit assignment during PPO updates we compute the log-probability from the original samples $\log\pi_{D}\!\bigl(\mathcal{C}^{(k)}\!\mid G^{(k)}\bigr)$.



Conditioned on the layout proposed by the design stage, the control stage optimizes signal timings to minimize pedestrian and vehicle delay. The control MDP evolves over $T$ action steps $t = 0, \dots, T-1$ within each SUMO episode. At each action step, the control policy sets a network-wide signal phase, i.e., the set of movements allowed at each controlled location. Each action persists for $R = 10$ simulation steps before the policy can act again.
\begin{table}[t]
\centering
\caption{Control agent action space. The agent selects from four phase configurations at the intersection and two at each mid-block crosswalk. Green denotes permitted movement; Red denotes prohibited. N, S, E, W denote cardinal directions.}
\label{tab:action-space}
\vspace{-6pt}
\setlength{\tabcolsep}{8pt}
\renewcommand{\arraystretch}{1.15}
\begin{tabular}{cccc}
\toprule
\textbf{Location} & \textbf{Phase} & \textbf{Vehicle} & \textbf{Pedestrian} \\
\midrule
\multirow{4}{*}{Intersection}
  & 1 & N-S Green, E-W Red   & E-W Green, N-S Red \\
  & 2 & E-W Green, N-S Red   & N-S Green, E-W Red \\
  & 3 & N-E Green, S-W Green & All Red \\
  & 4 & All Red               & All Green \\
\midrule
\multirow{2}{*}{Mid-Block}
  & 1 & Green & Red \\
  & 2 & Red   & Green \\
\bottomrule
\end{tabular}
\vspace{8pt}
\end{table}


\begin{itemize}

    \item \textbf{State.} The state $s_t$ is a two-dimensional spatio-temporal matrix encoding perceived traffic conditions across all controlled signals. Each action step $t$ spans $R$ simulation steps; the state is constructed by stacking: $s_t = \bigoplus_{j=0}^{R-1} \left[ \phi_{j}, v_{j}, p_{j} \right]$, where $\phi_{j}$ is the cumulative signal phase at simulation step $j$ across the network, $v_{j}$ is vehicle occupancy detected within $100$\,m for intersections and $50$\,m for crosswalks, categorized by incoming, inside, and outgoing lanes, and $p_{j}$ is pedestrian occupancy sensed within $5$\,m of crosswalks, categorized by incoming and outgoing directions. These detection ranges reflect typical vision-based intersection control systems~\cite{kong2024wts}.
    
    \item \textbf{Action.} The action $a_t$ specifies signal phases for all controlled signals, as summarized in Table~\ref{tab:action-space}. At the intersection, the policy selects from four phase configurations, each permitting a nonconflicting combination of vehicle and pedestrian movements, e.g., phase 1 allows N-S vehicle flow while pedestrians cross on the E-W sides. Right turns are always permitted, and left turns follow the straight-through green. At mid-block crosswalks, a binary action permits either vehicle flow or pedestrian crossing. The combined action space is $4 \times 2^{|\mathcal{C}^{(k)}|}$, growing with the number of crosswalks proposed by the design agent. Phase transitions include a 4-step yellow interval followed by all-red clearance. Further details are in supplementary material~\ref{appen:control}.
   
    \item \textbf{Reward.} To prevent a policy that optimizes average flow while inducing extreme worst-case delays, we build on the Maximum Wait Aggregated Queue (MWAQ)~\cite{koohy2022reward}, which couples the worst individual delay with accumulated demand. For a location with directions $D$ and waiting road users $V$ (pedestrians or vehicles), the baseline form is $\text{MWAQ} = \bigl(\max_{i\in V}\tau_i\bigr)\bigl(\sum_{d\in D} q_d\bigr)$, where $\tau_i$ is the waiting time of road user $i$ and $q_d$ is the queue length in direction $d$. We construct the control reward in three steps: (i) compute separate vehicle and pedestrian MWAQ terms at the intersection and at each crosswalk, (ii) aggregate crosswalk-level terms via the $L_2$ norm, and (iii) apply exponential penalties to suppress extreme delays. 

    \begin{figure}[t!] 
    \centering
    \includegraphics[width=0.99\linewidth]{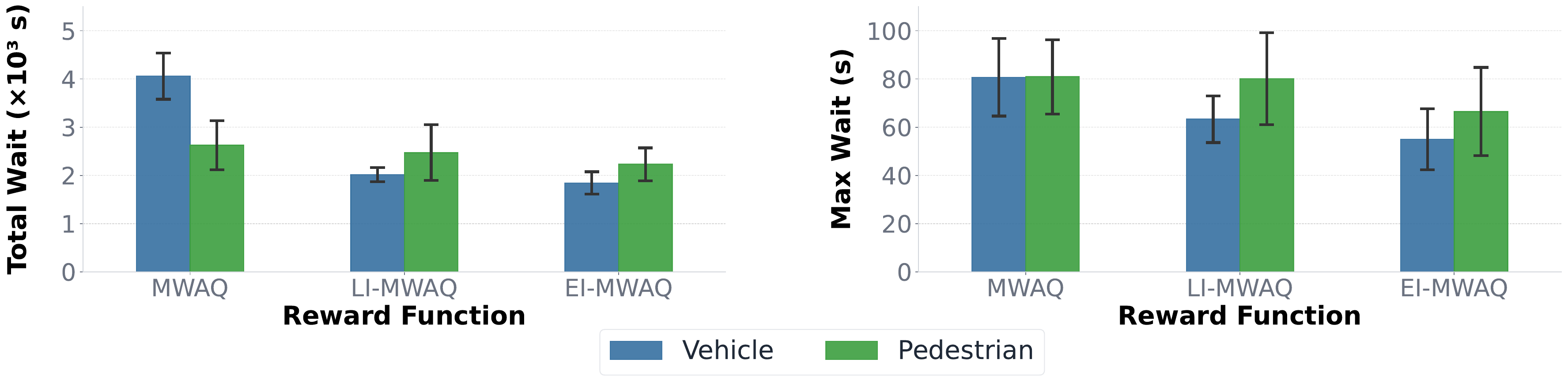}
    \vspace{-4pt}
    \caption{Effect of control reward on wait times: MWAQ, linearly increasing (LI-MWAQ), and exponentially increasing (EI-MWAQ) variants. Bar heights show averages over \(10\) runs; error bars denote \(\pm 1\) standard deviation. \textbf{LEFT:} Relative to MWAQ, LI-MWAQ reduces total wait times by \(50.3\%\) for vehicles and \(5.9\%\) for pedestrians, while EI-MWAQ yields \(54.6\%\) and \(15.1\%\), respectively. \textbf{RIGHT:} LI-MWAQ lowers maximum wait time by \(21.6\%\) for vehicles with negligible change for pedestrians (\(1.0\%\)), while EI-MWAQ reduces it by \(31.8\%\) and \(17.8\%\), respectively. EI-MWAQ consistently outperforms LI-MWAQ across both metrics, and the larger reductions in total wait time indicate that the penalties improve wait times system-wide, not merely for worst-case individuals.}
    \label{fig:ablation}
    \end{figure}
    
    At the intersection, the vehicle and pedestrian terms are
    \\[5pt]
    \noindent\begin{minipage}{0.48\linewidth}
    \begin{equation*}
    Q^{\mathrm{int}}_{\mathrm{veh}} = \beta_1 \!\left[\max_{i \in V^{\mathrm{int}}_{\mathrm{veh}}}\tau_i \sum_{d \in D} q_d^{\mathrm{veh}}\right]\!,
    \end{equation*}
    \end{minipage}\hfill
    \begin{minipage}{0.48\linewidth}
    \begin{equation*}
    Q^{\mathrm{int}}_{\mathrm{ped}} = \beta_2 \!\left[\max_{j \in V^{\mathrm{int}}_{\mathrm{ped}}}\tau_j \sum_{d \in D} q_d^{\mathrm{ped}}\right]\!.
    \end{equation*}
    \end{minipage}
    \\[5pt]
    Similarly, for each crosswalk $c \in \mathcal{C}^{(k)}$,
    \\[5pt]
    \noindent\begin{minipage}{0.48\linewidth}
    \begin{equation*}
    Q^{\mathrm{mb}}_{\mathrm{veh}}(c) = \beta_3 \!\left[\max_{i \in V^{\mathrm{mb}}_{\mathrm{veh}}(c)}\tau_i \sum_{d \in D_{\mathrm{mb}}} q_d^{\mathrm{veh}}(c)\right]\!,
    \end{equation*}
    \end{minipage}\hfill
    \begin{minipage}{0.48\linewidth}
    \begin{equation*}
    Q^{\mathrm{mb}}_{\mathrm{ped}}(c) = \beta_4 \!\left[\max_{j \in V^{\mathrm{mb}}_{\mathrm{ped}}(c)}\tau_j \, q^{\mathrm{ped}}(c)\right]\!.
    \end{equation*}
    \end{minipage}
    \\[5pt]
    In both cases, $V$ denotes the set of waiting road users at the respective location, $\tau$ is the individual waiting time, $q_d$ is the queue length in direction $d$, and $\beta_1$--$\beta_4$ are weighting coefficients.
    Because the number of crosswalks varies across rounds, we aggregate crosswalk-level terms via the $L_2$ norm to maintain a consistent reward scale,
    \\[5pt]
    \noindent\begin{minipage}{0.48\linewidth}
    \begin{equation*}
    Q^{\mathrm{mb}}_{\mathrm{veh}} = \left\|\left(Q^{\mathrm{mb}}_{\mathrm{veh}}(c)\right)_{c=1}^{|\mathcal{C}^{(k)}|}\right\|_2,
    \end{equation*}
    \end{minipage}\hfill
    \begin{minipage}{0.48\linewidth}
    \begin{equation*}
    Q^{\mathrm{mb}}_{\mathrm{ped}} = \left\|\left(Q^{\mathrm{mb}}_{\mathrm{ped}}(c)\right)_{c=1}^{|\mathcal{C}^{(k)}|}\right\|_2.
    \end{equation*}
    \end{minipage}
    \\[5pt]
    \noindent and the final reward applies exponential penalties over all terms,
    \begin{equation}
    R = -\left(e^{\beta_5 Q^{\mathrm{int}}_{\mathrm{veh}}} + e^{\beta_5 Q^{\mathrm{int}}_{\mathrm{ped}}} + e^{\beta_5 Q^{\mathrm{mb}}_{\mathrm{veh}}} + e^{\beta_5 Q^{\mathrm{mb}}_{\mathrm{ped}}}\right).
    \label{eq:control_reward_final}
    \end{equation}
    
\end{itemize}



To isolate the contribution of the exponential penalty, we hold the crosswalk layout fixed and train the control policy with three reward variants: the MWAQ baseline, a linear penalty, and our exponential penalty (EI-MWAQ). As shown in Fig.~\ref{fig:ablation}, EI-MWAQ consistently outperforms both alternatives across total and maximum wait times for vehicles and pedestrians, confirming that the exponential form is the most effective variant, which we adopt for all subsequent experiments.


The two stages use different actor-critic architectures. The design policy operates on pedestrian networks whose graph structure changes across rounds as crosswalks are added or removed. We encode this graph with a Graph Attention Network v2 (GATv2)~\cite{brody2021attentive}, whose dynamic attention mechanism naturally accommodates irregular topologies while conditioning on edge attributes. The encoder consists of two GATv2 layers with $8$ and $1$ attention heads, respectively, and is shared between the design actor and critic. Because node count varies across rounds, we apply global sort pool~\cite{zhang2018end}, ranking nodes by mean embedding activation and retaining the top $32$. The resulting fixed-length vector branches into actor and critic heads. The control actor and critic are separate multilayer perceptrons, the actor outputting a categorical distribution over signal phases at the intersection and independent Bernoulli distributions for each crosswalk.


\begin{figure*}[t!]
    \centering
    \includegraphics[width=0.98\linewidth]{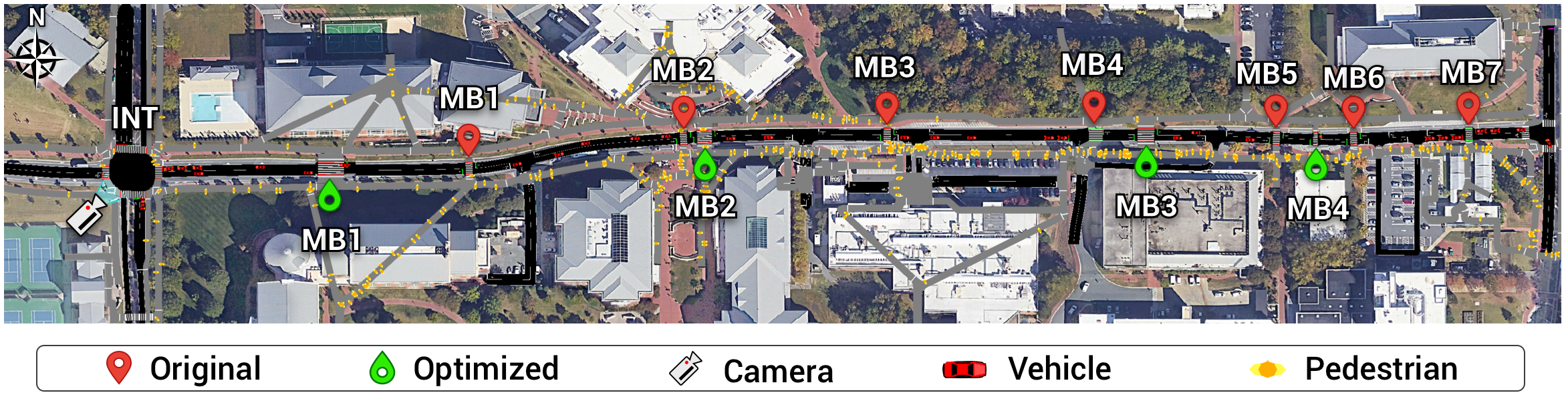}
    \vspace{-6pt}
    \caption{The real-world urban corridor before (red) and after (green) mid-block crosswalk layout optimization, with a demand of \(202\) veh/hr and \(2{,}223\) ped/hr. Vehicle demand is obtained from video data at intersection INT, and pedestrian demand from Wi-Fi logs. The existing mid-block crosswalks MB\(1\)--\(7\) are reduced to MB\(1\)--\(4\) proposed by \systemname{}, which better align with pedestrian desire lines and shorten walking paths.}
    \label{fig:craver}
\end{figure*}

We evaluate \systemname{} on a $750$\,m corridor on a university campus in the United States with over $34{,}000$ occupants across more than $80$ buildings, shown in Fig.~\ref{fig:craver}. The corridor includes one intersection (INT) at its western end and seven existing mid-block crosswalks (MB1--7). The intersection geometry remains fixed throughout the design stage, while the control stage regulates signal timings at the intersection and all mid-block crosswalks. Vehicle demand is obtained from video recordings at the intersection and pedestrian demand from $30$ days of anonymized Wi-Fi logs ($88{,}409$ unique clients across $2{,}492$ access points), yielding origin-destination flows of $202$ vehicles/hr and $2{,}223$ pedestrians/hr. When the crosswalk layout changes, the pedestrian router in the SUMO microsimulation recomputes shortest paths over the updated network without external adjustment~\cite{erdmann2015modelling}. Details of data collection, demand scaling, and re-routing are in supplementary material~\ref{appen:network_demand}.

%% file: sections/experiments.tex
\section{Experiment}
\subsection{Set-up}
\label{sec:setup}    

\begin{figure*}[t!]
    \centering
    \includegraphics[width=1.0\linewidth]{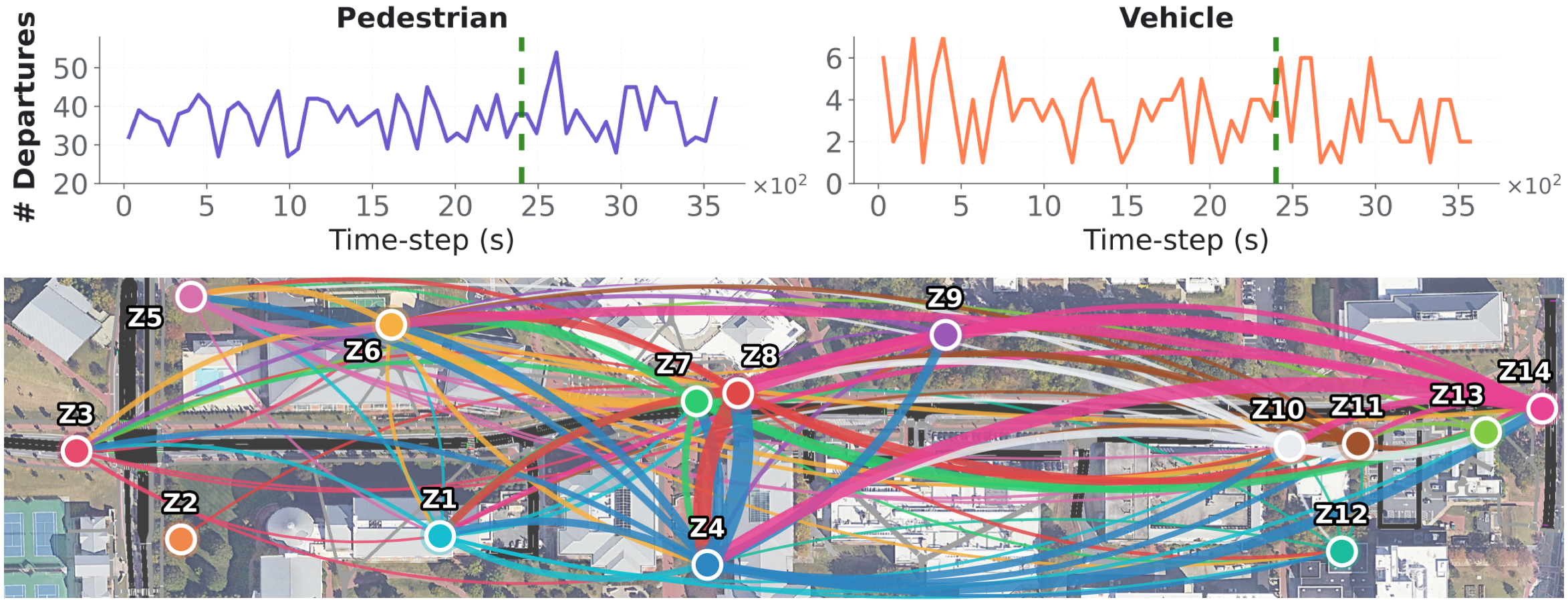}
    \vspace{-8pt}
    \caption{\textbf{TOP:} Real-world pedestrian (left) and vehicle (right) departure patterns obtained from data. The dashed line at \(t = 2{,}400\)\,s marks the train/evaluation split. Demand varies substantially between the two, ensuring distinct traffic conditions during training and evaluation. \textbf{BOTTOM:} Pedestrian origin-destination flow across \(14\) traffic analysis zones (Z1--Z14) in the study corridor; arcs represent flows between zones with thickness proportional to volume; \(69.6\%\) of trips (\(1{,}546\) of \(2{,}223\)) require crossing a road.}
    \vspace{2pt}
    \label{fig:craver_demand}
\end{figure*}

The demand data are split at $67\%$ ($2{,}400$\,s) for training, with the remaining $33\%$ ($1{,}200$\,s) held out for evaluation. As shown in Fig.~\ref{fig:craver_demand}, departure patterns differ substantially between the two periods. During training, demand is scaled uniformly in the range $[1.0, 2.25]$, while during evaluation we sweep $[0.5, 2.75]$ in increments of $0.25$, testing the policy on both unseen departure patterns and unseen scales. Each control episode runs across $10$ parallel environments, each with a different randomly scaled demand. Episodes are preceded by a warmup of $40$ to $140$ random steps, allowing pedestrians and vehicles to populate the network from their origins before the control policy acts. To evaluate the design policy in isolation from signal control, pedestrian arrival times under the optimized layout are compared against the existing configuration under an unsignalized scenario. On the control side, performance is evaluated against two benchmarks: an unsignalized baseline where pedestrians have right-of-way at mid-block crosswalks while the intersection remains signalized, and a fixed-time baseline with cycle lengths derived from traffic engineering guidelines~\cite{traffic2023manual}. More details on benchmarks are in supplementary material~\ref{appen:control_benchmarks}.


\subsection{Results}
\label{sec:results}
Our results demonstrate that \systemname{} improves both pedestrian and vehicle mobility. We evaluate performance across pedestrian arrival time to nearest mid-block crosswalk, network-wide pedestrian and vehicle wait times, generalization to unseen demands, and robustness of the control policy to layout changes.

\begin{figure*}[t!]
    \centering
    \vspace{-2pt}
    \includegraphics[width=0.99\linewidth]{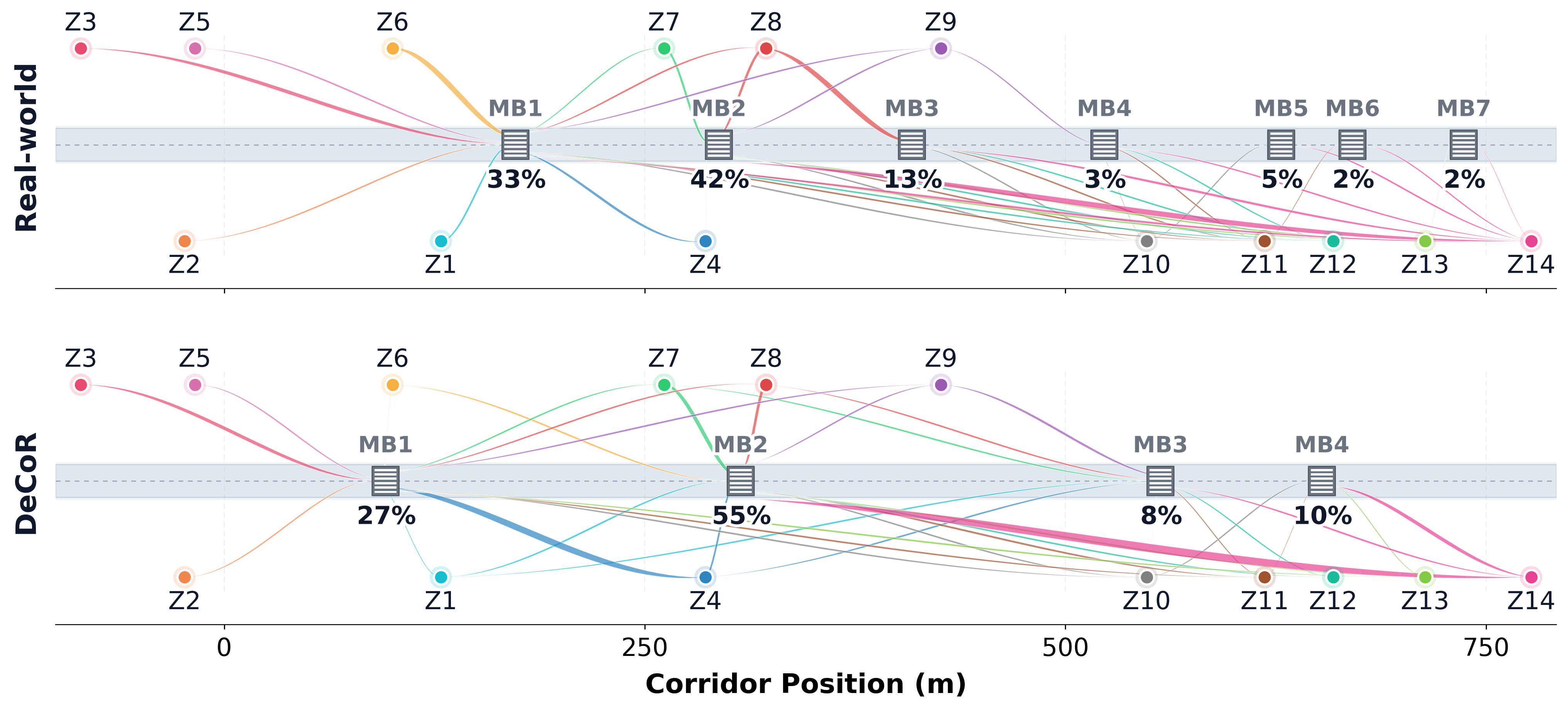}
    \vspace{-8pt}
    \caption{Pedestrian flow allocation under real-world and \systemname{} layouts at \(1\times\) demand. Arcs connect traffic analysis zones (Z1--Z14) to the crosswalk on each pedestrian's shortest-path route. Arc width is proportional to volume; only the origin-to-crosswalk segment is shown. \textbf{TOP:} The real-world layout uses seven crosswalks (MB1--7), where \(75\%\) of demand concentrates at MB1 (\(33\%\)) and MB2 (\(42\%\)), while MB5--7 each serve only \(2\text{--}5\%\). \textbf{BOTTOM:} \systemname{} reduces the count to four crosswalks (MB1--4), where MB1 absorbs \(27\%\) of demand from zones previously detouring east and MB2 consolidates the two busiest real-world locations into a single crosswalk serving \(55\%\). Together, MB1 and MB2 handle \(82\%\) of demand at fewer, better-positioned locations.}
    \label{fig:flow_allocation}
    \vspace{-6pt}
\end{figure*}

\begin{figure*}[t!]
    \centering
    \includegraphics[width=1.0\linewidth]{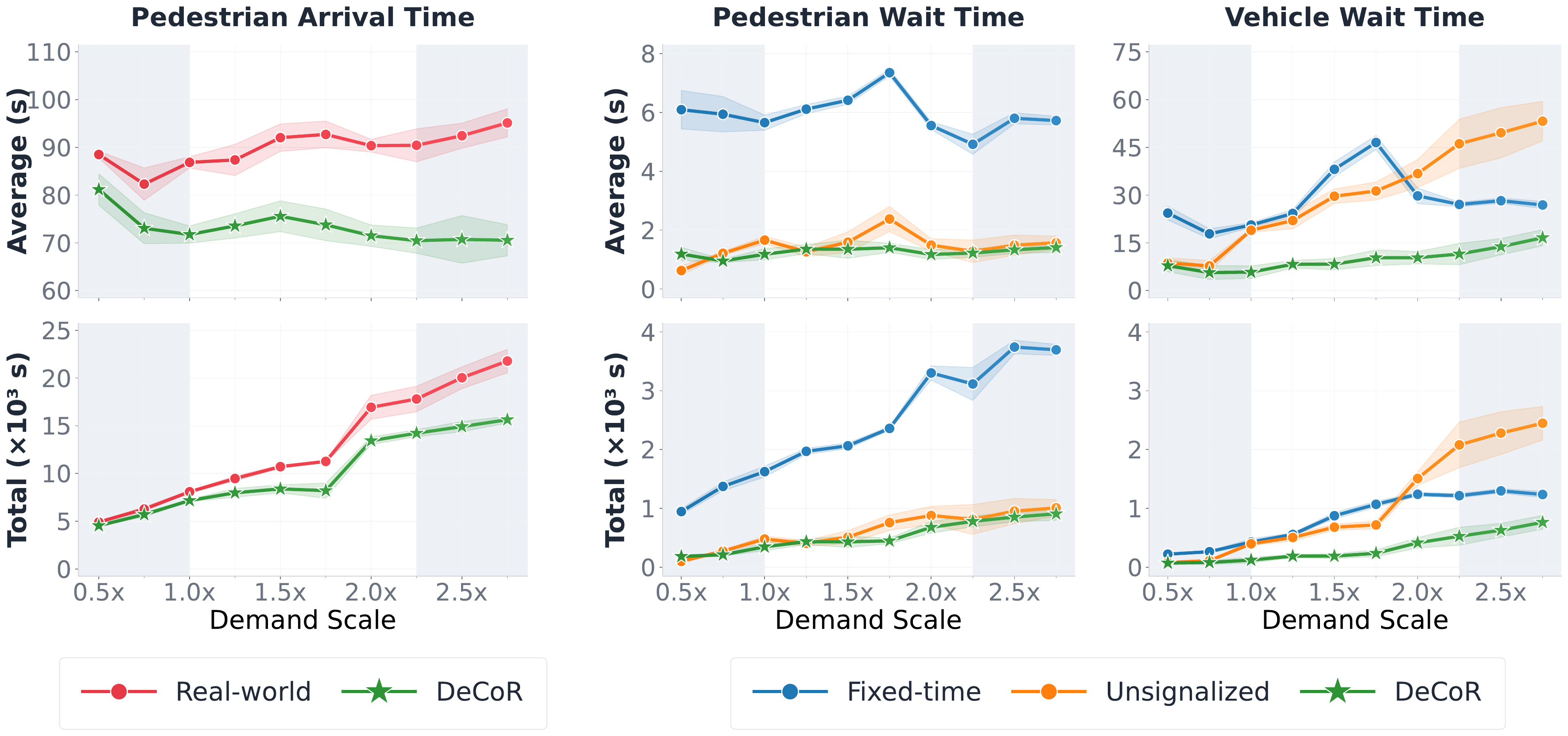}
    \vspace{-16pt}
    \caption{Travel time metrics across varying demands. \textbf{LEFT}: Pedestrian arrival times from \systemname{}'s \(4\) crosswalk placements against the real-world \(7\) crosswalk configuration, showing consistent reductions in both average and total arrival times. \textbf{MIDDLE}: Pedestrian wait times at crossings under different signalization strategies, showing that \systemname{}'s control (green) achieves wait times comparable to unsignalized (orange) and lower than fixed-time (blue) across all demands. \textbf{RIGHT}: Vehicle wait times showing that \systemname{}'s control reduces delays compared to fixed-time across all demands, with particularly significant improvements over unsignalized approaches at higher demands. Gray regions (\({<}1\times\) and \({>}2.25\times\) demand) indicate scales not encountered during training. Metrics are averaged over \(10\) runs; shaded regions show \(\pm 1\) standard deviation.}
    \vspace{2pt}
    \label{fig:main_results}
\end{figure*}

For pedestrian arrival time, the mid-block crosswalk configuration proposed by our design agent (\(4\) crosswalks) is compared against the existing real-world layout (\(7\) crosswalks) across varying pedestrian demand scales, as shown in Fig.~\ref{fig:main_results} LEFT. The optimized design consistently reduces pedestrian arrival times compared to the real-world configuration. Each pedestrian in the real-world layout who intends to cross the corridor takes \(89.82 \pm 4.37\)\,s to reach their nearest mid-block crosswalk versus \(73.21 \pm 4.42\)\,s in the optimized design. On average, this represents a \(22.68\%\) reduction in arrival time. Improvements are particularly pronounced at higher demand scales, e.g., at \(2.75\)× demand, the optimized design reduces average arrival time from \(95.13 \pm 2.95\)\,s to \(70.55 \pm 3.27\)\,s, a \(25.8\%\) improvement. The pedestrian flow allocation in Fig.~\ref{fig:flow_allocation} shows that \systemname{} aligns crosswalks with desire lines, with MB1 and MB2 alone handling \(82\%\) of demand at fewer and better-positioned locations, shortening detours even as the total count decreases.

For pedestrian wait time, the traditionally held notion that unsignalized crossings are fast while signalized crossings are safe but slow is challenged by our control agent. Fig.~\ref{fig:main_results} MIDDLE compares our agent against unsignalized and fixed-time signalized benchmarks across varying pedestrian demand scales. Under our control agent, each pedestrian experiences a wait time of \(1.26 \pm 0.23\,\text{s}\), a \(78.9\%\) reduction compared to fixed-time signalization (\(5.95 \pm 0.69\,\text{s}\)). This closely matches the unsignalized scenario (\(1.46 \pm 0.50\,\text{s}\)), often considered the lower bound for pedestrian delay. Even as pedestrian demand increases, our control agent consistently maintains efficient crossing opportunities without compromising safety (no pedestrian-vehicle conflicts).

For vehicle wait time, as shown in Fig.~\ref{fig:main_results} RIGHT, each vehicle experiences a wait time of \(9.85 \pm 4.00\)\,s under our control agent, a \(65.3\%\) reduction compared to fixed-time signalization (\(28.37 \pm 8.18\)\,s) and a \(67.6\%\) reduction compared to the unsignalized scenario (\(30.41 \pm 16.11\)\,s). A crossover pattern emerges across demand scales: at lower demand, unsignalized crossings yield shorter vehicle wait times than fixed-time signals; however, as demand increases, vehicle wait times in the unsignalized scenario rise sharply, eventually surpassing fixed-time signalization and reaching \(53.23 \pm 6.53\)\,s at \(2.75\times\) demand. This demonstrates the limitation of unsignalized crossings at higher demands.

To test generalization, we evaluate both agents on demand scales outside the training distribution, including low-demand (\(0.50\times\), \(0.75\times\)) and high-demand (\(2.50\times\), \(2.75\times\)) scenarios, indicated by the gray-shaded regions in Fig.~\ref{fig:main_results}. The design agent's optimized crosswalk configuration maintains its advantage over the real-world layout at both extremes, with a \(25.8\%\) improvement in average arrival time per pedestrian at \(2.75\times\) demand. For the control agent, at very low demand (\(0.5\times\)), improvements over unsignalized control are modest (\(9.8\%\) per pedestrian on average), as unsignalized crossings naturally perform well in low-traffic conditions. At very high demand (\(2.75\times\)), the control agent outperforms fixed-time signalization by \(38.2\%\) per pedestrian on average. Across all scenarios, the control agent achieves this without a single pedestrian--vehicle conflict. Both agents effectively generalize to unseen demand conditions.

\begin{figure*}[t!]
    \centering
    \includegraphics[width=0.99\linewidth]{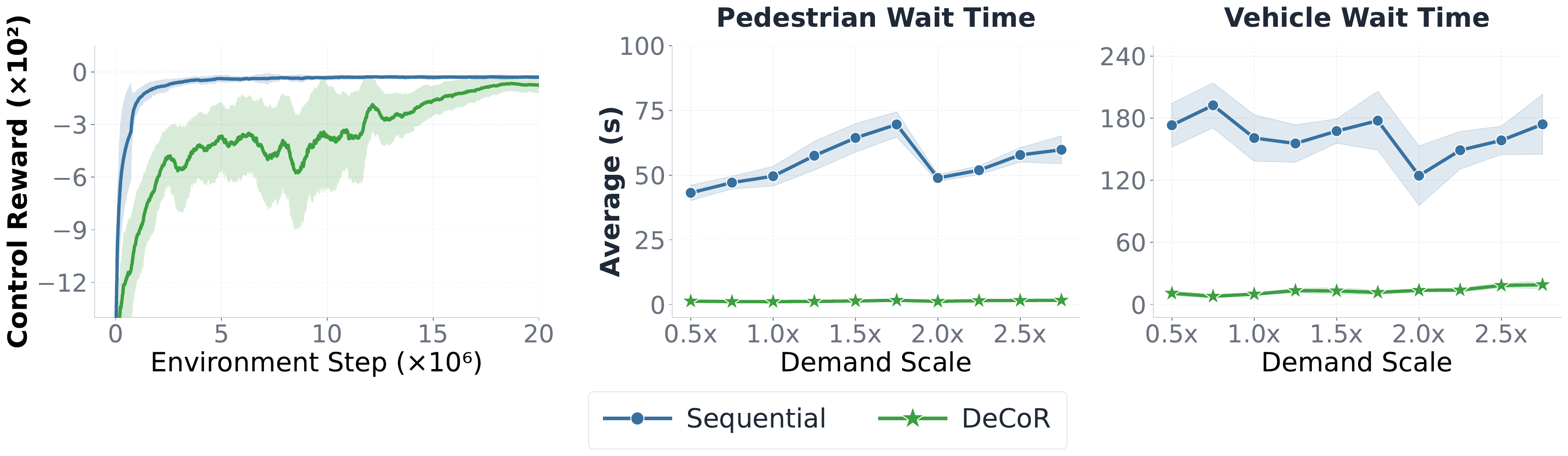}
    \vspace{-4pt}
    \caption{\textbf{LEFT:} Control agent reward during training under \systemname{} (co-optimization) versus sequential training, i.e., design first, then control. The raw rewards average \(-76.8 \pm 48.5\) for \systemname{} and \(-28.8 \pm 12.1\) for sequential over the last \(5 \times 10^{5}\) steps. Values are averaged over three random seeds with shaded regions denoting \(\pm 1\) standard deviation. Despite lower training reward from facing varying layouts, wall-clock convergence is comparable to sequential training; yet the resulting policy is far more robust. \textbf{MIDDLE, RIGHT:} Pedestrian and vehicle wait times of both approaches when evaluated with an additional \(5\)\,m wide crosswalk placed on the optimized four-crosswalk layout. Across demand scales, \systemname{}'s control agent maintains lower wait times than the sequentially trained controller, with up to \(97.53\%\) lower pedestrian and up to \(91.90\%\) lower vehicle waiting times. Results are averaged over \(10\) runs.}
    \label{fig:second_results}
\end{figure*}

Co-optimization exposes the control policy to a distribution of layouts as the design policy proposes different crossing configurations during training. In contrast, a sequential alternative that first optimizes the layout and then trains the controller on that fixed geometry only ever sees one network. To test robustness to structural changes at test time, we add a fifth mid-block crosswalk of width \(5\)\,m at the east end of the optimized four-crosswalk layout, and evaluate both controllers without retraining. As shown in Fig.~\ref{fig:second_results} MIDDLE and RIGHT, the co-optimized controller maintains low delay under this layout shift, while the sequential controller degrades dramatically. Averaged over demand scales, the co-optimized controller yields \(1.36 \pm 0.21\)\,s pedestrian wait and \(13.23 \pm 2.13\)\,s vehicle wait, while the sequential controller reaches \(55.06\)\,s and \(163.33 \pm 21.31\)\,s, respectively, a \(97.53\%\) and \(91.90\%\) difference. For reference, the original four-crosswalk layout without the added crosswalk yields \(1.25 \pm 0.22\)\,s pedestrian wait and \(10.06 \pm 3.98\)\,s vehicle wait, indicating that the co-optimized controller barely degrades despite the structural modification. Together, evaluation on unseen demand patterns, unseen demand scales, and unseen layout changes demonstrates generalization of both agents within a single corridor.\

Notably, the robustness gained through co-optimization comes at no meaningful efficiency cost. Although the design stage introduces approximately $10$\% computational overhead per round from graph-to-simulation conversion, total wall-clock training time is $36$ hours.

%% file: sections/conclusion.tex
\section{Discussion and Conclusion}
\label{sec:conclusion}

In this paper, we ask whether the gap from \emph{perception to design} can be closed for urban pedestrian mobility, and whether sensing how pedestrians and vehicles move through a corridor is sufficient to jointly redesign the crossing layout and adapt its control. To answer this question, we introduce \systemname{}, a two-stage reinforcement learning framework that co-optimizes two coupled decisions shaping pedestrian mobility: \emph{where to place crosswalks} and \emph{how to regulate them}, grounding both in demand sensed from video and Wi-Fi data and scoring both through closed-loop simulation of pedestrian and vehicle travel times. On a real-world corridor, the layout designed by \systemname{} uses four mid-block crosswalks instead of the existing seven and reduces pedestrian arrival time by up to 23\%, while the control policy cuts pedestrian and vehicle wait time by up to 79\% and 65\%, respectively, relative to fixed-time signalization. Both policies generalize to unseen demand patterns and scales, and the co-optimized controller remains robust to structural layout changes without retraining, achieving up to 97\% lower wait times than a sequentially trained counterpart.

A notable outcome of co-optimization is that a \emph{pedestrian-first design} reduces pedestrian arrival time despite using fewer crossings, and simultaneously reduces vehicle delay. This occurs because aligning crosswalks with pedestrian desire lines shortens detours even as the total count decreases, while consolidating crossings into fewer, well-placed locations paired with adaptive signal timing reduces stop-and-go burden on vehicles. This \emph{joint improvement} is precisely what isolated design or isolated control cannot reliably achieve. Critically, co-optimization introduces no meaningful training overhead relative to sequential training, yet delivers these benefits. These results also challenge the common assumption that signalized crossings are inevitably slow for pedestrians: with perception-based state feedback, adaptive control can approach the convenience of unsignalized crossings while maintaining vehicle efficiency under high demand.

At the same time, we emphasize the scope of what is claimed. Our design objective targets mobility, using pedestrian arrival time as a surrogate for reducing detours that can motivate unsafe mid-block crossings, but \systemname{} does not explicitly model injury risk or fatalities. Likewise, our evaluation focuses on delay and conflict-free operation in simulation, and does not model signal violations or surrogate safety metrics such as time-to-collision. In simulation, we assume access to pedestrian and vehicle state observations, whereas real deployments rely on sensing systems subject to noise, occlusion, and missing data. A practical limitation is that we evaluate on a single real-world corridor rather than multiple unrelated networks, driven largely by the scarcity of publicly available datasets that pair realistic network geometry with high-fidelity, temporally aligned pedestrian and vehicle demand measurements for the same site. We partially address generalization within the available data by testing under temporally held-out demand patterns, demand scaling beyond the training range, and robustness to a structural layout change at test time without retraining.

Future work includes extending the design space beyond the one-dimensional coordinate imposed by the single-corridor setting to higher-dimensional network geometries, which motivates relaxing the GMM's diagonal covariance and fixed standard deviation to capture richer spatial correlations. Incorporating conflict-based indicators such as post-encroachment time and modeling explicit noncompliance are additional directions. Transferring policies across networks without adaptation also remains an open direction. \systemname{} establishes that the gap from perception to design can be closed, and that sensing how pedestrians and vehicles move through a corridor is sufficient to design the infrastructure that shapes their behavior.

%% file: sections/acknowledgements.tex
\subsection*{Acknowledgements}
The authors would like to thank Jakob Erdmann of SUMO for helping with technical issues in simulation, Dr. Christopher R. Cherry of the Department of Civil and Environmental Engineering at the University of Tennessee for motivating the direction of the project, and Chandra Raskoti for proofreading the work. 

%% file: sections/supplementary.tex
\section*{Supplementary Material}

\section{Design Agent}
\label{appen:policy_gmm}

We adopt a GMM-based policy parameterization because GMMs naturally capture multimodal distributions, enabling the policy to explore multiple promising crosswalk configurations across the corridor. GMMs also support differentiable sampling via the reparameterization trick: the policy network outputs the GMM means and configurations are sampled from the learned distribution while preserving gradient flow. The GMM framework separates exploration from exploitation: during training, we sample stochastically to explore diverse configurations; during evaluation, we deterministically select modes by identifying local maxima of the learned density, yielding interpretable and consistent designs.

We fix the diagonal covariance at $\sigma = e^{-2.5}$ with crosswalk location in the range $[2\,\text{m},\, 748\,\text{m}]$ and width in the range $[2\,\text{m},\, 15\,\text{m}]$. Learning full covariance matrices results in quadratic growth $O(Md^2)$ in the number of free parameters, whereas learning only the component means results in linear growth $O(Md)$. In words, restricting learnable parameters to the means reduces the optimization surface from quadratic to linear in the number of mixture components, improving stability while retaining capacity for distinct crosswalk placements. Extending to higher-dimensional network geometries would motivate learning full covariance matrices to capture correlations between crosswalk location and width, and learnable mixture weights to distribute attention across corridor regions.

\section{Control Agent}
\label{appen:control}

\subsection{Action Space}

Permitted-turn behavior is governed by the simulation platform. Similar nonconflicting phase sets can be derived for other intersection geometries, making the formulation broadly applicable.

\subsection{Reward Scalars and Normalization}
\label{appen:reward_scalars}

The control reward uses empirically chosen scaling constants: $\beta_1 = 1/(2|D_{\mathrm{int}}|)$, $\beta_2 = 1/(10|D_{\mathrm{int}}|)$, $\beta_3 = 1/(2|D_{\mathrm{mb}}|)$, $\beta_4 = 1/10$, and $\beta_5 = 0.5$, where $D_{\mathrm{int}}$ and $D_{\mathrm{mb}}$ denote the number of incoming directions at intersections and mid-block crosswalks, respectively. The denominators $2$ (for vehicles) and $10$ (for pedestrians) balance MWAQ magnitudes, reflecting that pedestrian queue counts are typically an order of magnitude larger than vehicle counts. Crosswalk-level terms are aggregated via the $L_2$ norm rather than a sum or maximum: the $L_2$ norm scales sub-linearly with the number of crosswalks, preventing reward magnitude from growing proportionally as crossings are added while penalizing extreme individual delays more heavily than a simple average.

Raw rewards are clipped to $[-2500,\,0]$. For numerical stability, states and rewards are normalized via Welford's algorithm~\cite{huang202237,rl_bag_of_tricks}. Vehicles are considered waiting when their speed falls below $0.2\,\mathrm{m/s}$, and pedestrians below $0.5\,\mathrm{m/s}$.

\subsection{Benchmarks}
\label{appen:control_benchmarks}

The control agent is evaluated against two baselines. The first is an unsignalized configuration in which all mid-block locations operate as unsignalized crossings, matching the corridor's current real-world setup. Pedestrians have the right-of-way as specified by the Uniform Vehicle Code~\cite{UVC2000}, implemented using SUMO's pedestrian interaction model~\cite{sumo_pedestrians}, where vehicles yield to pedestrians at shared roadway segments and designated crossings. This pedestrian-prioritized baseline minimizes pedestrian delay at mid-block locations but may increase vehicle delay.

The second baseline implements fixed-time signal control at both the intersection and mid-block crosswalks, with timings derived from real-world observations and standard traffic engineering guidelines:

\begin{itemize}
    \item \textbf{Intersection:} Operates on a five-phase cycle with $90$-second green intervals alternating between north-south (N-S) and east-west (E-W) through vehicle movements. Each directional change includes a $4$-second yellow interval followed by a $2$-second all-red clearance period. Consistent with the current real-world implementation, the signal timing does not include dedicated left-turn phases. Signalized pedestrian crossings at the intersection activate simultaneously with their corresponding vehicle phases; specifically, when the north-south vehicle movement is green, the non-conflicting pedestrian crosswalks oriented east-west are also green, and vice versa. These timings are determined through manual observation of video footage captured at various times of the day.
    \item \textbf{Mid-block crosswalks:} Operate on a $62$-second fixed cycle, with phases set according to guidelines from the Federal Highway Administration's Manual on Uniform Traffic Control Devices (MUTCD)~\cite{traffic2023manual} and the Traffic Signal Timing Manual (TSTM)~\cite{koonce2008traffic}. The pedestrian phase (MUTCD 4I$.06$) lasts $16$ seconds, consisting of a $7$-second minimum walk interval followed by a $9$-second pedestrian clearance interval, calculated as:
    \[
        \text{Clearance time} = \frac{\text{Crosswalk length}}{\text{Walking speed}} = \frac{32 \text{ ft}}{3.5 \text{ ft/s}} \approx 9 \text{ s}.
    \]
    The vehicle phase (TSTM $6.6.3$, MUTCD 4F$.17$) lasts $46$ seconds, comprising a $40$-second green interval (approximately $64\%$ of the cycle), a $4$-second yellow change interval, and a $2$-second red clearance interval.
\end{itemize}

The signalized benchmark represents a fully controlled baseline with dedicated phases designed to eliminate vehicle-pedestrian conflicts. The control agent replaces these fixed cycles with adaptive timings regulated by the learned policy.

\section{Network, Demand, and Simulation}
\label{appen:network_demand}

\subsection{Data Collection and Processing}

Wi-Fi logs for pedestrian demand estimation are collected by the university's IT department and fully anonymized; no personally identifiable information is retained. An anonymized version is part of the released materials.

Pedestrian demand is extracted from anonymized Wi-Fi logs collected in September $2021$ across $2{,}492$ access points with $88{,}409$ unique clients. Each entry records when and where a client connects. Three processing steps are applied:
\begin{itemize}
    \item Client activities are aggregated at the building level: sessions within the same building are merged to represent presence at that location.
    \item Clients detected on fewer than three days per month ($11.81\%$ of the dataset) are classified as visitors or irregular commuters and excluded, as the analysis targets typical campus travel patterns.
    \item To address individuals carrying multiple devices, we apply K-means clustering to classify and remove non-mobile devices based on the mean and variance of their stationary ratio relative to total daily activity time~\cite{zhang2023crowdtelescope}.
\end{itemize}

Vehicle demand is estimated from four video recordings (total $21$ minutes) captured at different times of day at the intersection. Vehicle counts are extrapolated to an hourly rate with random perturbations applied to departure times, resulting in an average headway of $18$ seconds. Both pedestrian and vehicle streams are mapped to SUMO~\cite{krajzewicz2002sumo} trip definitions. Origin-destination pairs are defined using Traffic Analysis Zones: pedestrian pairs from building visit logs, vehicle pairs from recorded turning movements.

\subsection{Demand Scaling and Routing}
\label{app:sample}

To simulate varying traffic loads, we scale each episode by a factor $\alpha_i$. First, we compress the episode timeline via $t' = (t - t_{\mathrm{start}})/\alpha_i$ for $0 \le t - t_{\mathrm{start}} < T$, shortening the episode to $T/\alpha_i$ and increasing the instantaneous departure rate by $\alpha_i$. Second, we replicate the compressed trips $\alpha_i$ times and distribute them evenly across the original duration, preserving total demand. In words, the compression concentrates the original departures into a shorter window to raise the instantaneous rate, while replication fills the remaining episode duration so that the total number of trips scales proportionally with $\alpha_i$. Time-window sampling and demand scaling are standard practices for tractability and sample efficiency, with built-in support in SUMO~\cite{krajzewicz2002sumo} and CityFlow~\cite{zhang2019cityflow}.

Since demand data specifies origin-destination pairs rather than fixed routes, SUMO automatically computes the shortest path for each pedestrian trip given the current layout~\cite{erdmann2015modelling,SUMO2018}. When the crosswalk configuration changes, pedestrians dynamically re-route without manual adjustment, so each design iteration is evaluated by updating the network description. Pedestrians still obey traffic signals along their chosen path.

\section{Training Configuration}
\label{app:hyperparams}

Training uses $10$ parallel lower-level control actors over $20\times10^{6}$ total simulation steps on an Intel Core i$9$-$14900$KF processor and an NVIDIA RTX $6000$ PRO GPU. Total wall-clock training time is $36$ hours. Table~\ref{table:all_params} summarizes the simulation settings, architecture, and PPO hyperparameters for both agents.

\begin{table}[t!]
    \centering
    \setlength{\tabcolsep}{4pt}
    \renewcommand{\arraystretch}{1.05}
    \caption{Simulation, architecture, and PPO hyperparameters for both agents.}
    \begin{tabular}{clcc}
        \toprule
        & \textbf{Parameter} & \textbf{Design} & \textbf{Control} \\
        \midrule
        \multirow{6}{*}{\rotatebox{90}{\textbf{Simulation}}}
        & Simulation Step (\(R\)) & \multicolumn{2}{c}{\(0.1\)\,s} \\
        & Action Step (\(\Delta t\)) & \multicolumn{2}{c}{\(1\)\,s} \\
        & Episode Horizon (\(T\)) & \multicolumn{2}{c}{\(360\)} \\
        & Warmup Steps & \multicolumn{2}{c}{\([40, 140]\)} \\
        & Total Sim.\ Steps & \multicolumn{2}{c}{\(20\!\times\!10^{6}\)} \\
        & Location Range & \multicolumn{2}{c}{\([2, 748]\)\,m} \\
        & Width Range & \multicolumn{2}{c}{\([2, 15]\)\,m} \\
        \midrule
        \multirow{9}{*}{\rotatebox{90}{\textbf{Architecture}}}
        & Encoder & GATv2 (2 layers) & MLP \\
        & Attn.\ Heads & (8, 1) & -- \\
        & Input Channels (\(F_e\)) & \(2\) & -- \\
        & Hidden / Out Dims & \(64\) / \(64\) & -- \\
        & Sort-pool (\(k\)) & \(32\) & -- \\
        & Shared MLP & 512, 256 & -- \\
        & Actor MLP & 256, 128, 64 & 512, 256, 128, 64 \\
        & Critic MLP & 256, 128, 64 & 512, 256, 128, 64 \\
        & Activation & tanh & tanh \\
        \midrule
        \multirow{11}{*}{\rotatebox{90}{\textbf{PPO}}}
        & Learning Rate & \(5\!\times\!10^{-4}\) & \(5\!\times\!10^{-4}\) \\
        & Anneal LR & True & False \\
        & GAE \(\lambda\) & \(0.97\) & \(0.95\) \\
        & Discount \(\gamma\) & \(0.99\) & \(0.99\) \\
        & Epochs & \(4\) & \(2\) \\
        & Clip \(\epsilon\) & \(0.3\) & \(0.1\) \\
        & Entropy Coeff. & \(0.001\) & \(0.005\) \\
        & VF Coeff. & \(0.5\) & \(0.5\) \\
        & Batch Size & \(2\) & \(32\) \\
        & Update Freq. & \(16\) & \(1024\) \\
        & \# Parallel Actors & -- & \(10\) \\
        \bottomrule
    \end{tabular}
    \label{table:all_params}
\end{table}